%% file: root.tex
\documentclass[letterpaper, 10 pt, conference]{ieeeconf}  

\IEEEoverridecommandlockouts                              

\usepackage[dvipsnames]{xcolor}

\usepackage{amsthm}
\usepackage{amsmath}
\usepackage[font=scriptsize,labelfont=bf]{caption}
\usepackage[font=scriptsize,labelfont=bf]{subcaption}
\usepackage{float}
\usepackage{algorithm}
\usepackage{algpseudocode}
\usepackage{multicol}
\usepackage{duckuments}
\usepackage{graphicx}
\usepackage{xr}
\let\labelindent\relax
\usepackage{enumitem}
\usepackage{etoolbox}
\usepackage{mathrsfs}
\usepackage{multirow}
\usepackage{balance}
\usepackage{overpic}
\usepackage{nccmath}
\usepackage{titlesec}
\usepackage[nomath]{cellspace}

\DeclareCaptionType{CapEq}[][]
\definecolor{tab:blue}{RGB}{31,119,180}
\definecolor{tab:orange}{RGB}{255,127,14}
\definecolor{tab:green}{RGB}{44,160,44}
\definecolor{tab:red}{RGB}{214,39,40}
\definecolor{tab:purple}{RGB}{148,103,189}

\definecolor{sns:blue}{rgb}{0.00392156862745098, 0.45098039215686275, 0.6980392156862745}
\definecolor{sns:orange}{rgb}{0.8705882352941177, 0.5607843137254902, 0.0196078431372549}
\definecolor{sns:green}{rgb}{0.00784313725490196, 0.6196078431372549, 0.45098039215686275}
\definecolor{sns:red}{rgb}{0.8352941176470589, 0.3686274509803922, 0.0}
\definecolor{sns:purple}{rgb}{0.8, 0.47058823529411764, 0.7372549019607844}
\definecolor{sns:brown}{rgb}{0.792156862745098, 0.5686274509803921, 0.3803921568627451}
\definecolor{sns:pink}{rgb}{0.984313725490196, 0.6862745098039216, 0.8941176470588236}
\definecolor{sns:gray}{rgb}{0.5803921568627451, 0.5803921568627451, 0.5803921568627451}
\definecolor{sns:yellow}{rgb}{0.9254901960784314, 0.8823529411764706, 0.2}
\definecolor{sns:light_blue}{rgb}{0.33725490196078434, 0.7058823529411765, 0.9137254901960784}
\newtheoremstyle{dense}
  {3pt} 
  {3pt} 
  {\itshape} 
  {} 
  {\bfseries} 
  {:} 
  {.5em} 
  {} 
\theoremstyle{dense}

\newtheorem{remark}{Remark}
\AtBeginEnvironment{remark}{\small}
\AtBeginEnvironment{quote}{\par\singlespacing\small}
\AtBeginEnvironment{tabular}{\footnotesize}
\IEEEaftertitletext{\vspace{-1.4\baselineskip}}
\titleformat{\subsubsection}[runin]
   {\itshape}
   {\thesubsubsectiondis}
   {0.5em}
   {}
   [:]
\titlespacing\subsubsection{\parindent}{0pt}{0.5em}
\titlespacing\subsection{0pt}{4pt plus 1pt minus 1pt}{2pt plus 1pt minus 1pt}
\titlespacing\section{0pt}{6pt plus 1pt minus 1pt}{2pt plus 1pt minus 1pt}
\usepackage{stix}
\usepackage{tikz}
\DeclareRobustCommand{\thickX}{
    \begin{tikzpicture}[baseline=0ex, line width=2, scale=0.13];
    \draw (0,0) -- (1,1);
    \draw (0,1) -- (1,0);
    \end{tikzpicture}}
\DeclareRobustCommand{\thickY}{
    \begin{tikzpicture}[baseline=0ex, line width=0.75, scale=0.15];
    \draw (0.5,0) -- (0.5,0.5);
    \draw (0,1) -- (0.5,0.5);
    \draw (1,1) -- (0.5,0.5);
    \end{tikzpicture}}
\DeclareRobustCommand{\thickPlus}{
    \begin{tikzpicture}[baseline=0ex, line width=1.75, scale=0.15];
    \draw (0.5,0) -- (0.5,1);
    \draw (0,0.5) -- (1,0.5);
    \end{tikzpicture}}
\newcommand{\SymMesaGeo}{\textcolor{sns:blue}{$\mdblkcircle$}}
\newcommand{\SymMesaSplit}{\textcolor{sns:green}{$\mdblksquare$}}
\newcommand{\SymMesaOneOrd}{\textcolor{sns:yellow}{$\blacktriangle$}}
\newcommand{\SymMesaManif}{\textcolor{sns:pink}{$\thickY$}}
\newcommand{\SymDgs}{\textcolor{sns:red}{$\thickPlus$}}
\newcommand{\SymMbAdmm}{\textcolor{sns:purple}{$\blacklozenge$}}
\newcommand{\SymAsapp}{\textcolor{sns:brown}{$\thickX$}}
\newcommand{\SymCent}{\textcolor{sns:gray}{$\bigstar$}}
\newcommand{\SymIndep}{\textcolor{sns:orange}{$\mathrm{\textit{\textbf{I}}}$}} 
\newcommand{\SymDdf}{\textcolor{sns:light_blue}{$\pentagonblack$}}
\makeatletter
\let\NAT@parse\undefined
\makeatother
\usepackage{hyperref}
\usepackage{cleveref}
\hypersetup{
    colorlinks=true,
    linkcolor=blue,
    filecolor=magenta,      
    urlcolor=cyan,
    pdfpagemode=FullScreen,
}

\title{\LARGE \bf Asynchronous Distributed Smoothing and Mapping via On-Manifold Consensus ADMM \vspace{-2ex}
\author{Daniel McGann$^1$, Kyle Lassak$^2$, and Michael Kaess$^1$}
\thanks{\scriptsize
This work was partially supported by NASA award 80NSSC22PA952, the NSF Graduate Research Fellowship Program, and DEVCOM Army Research Laboratory Distributed, Collaborative, Intelligent Systems and Technology Collaborative Research Alliance (DCIST-CRA) W911NF-17-2-0181.}
\thanks{ \scriptsize
$^1$ D. McGann and M. Kaess are with the Robotics Institute, Carnegie Mellon University, Pittsburgh, PA, USA. \texttt{\{danmcgann, kaess\}@cmu.edu}
}
\thanks{ \scriptsize
$^2$ K. Lassak is with Astrobotic Technology Inc., Pittsburgh, PA, USA, \texttt{kyle.lassak@astrobotic.com}
}
}

\input{helpers}

\begin{document}

\maketitle
\thispagestyle{empty}
\pagestyle{empty}

\begin{abstract}
In this paper we present a fully distributed, asynchronous, and general purpose optimization algorithm for Consensus Simultaneous Localization and Mapping (CSLAM). Multi-robot teams require that agents have timely and accurate solutions to their state as well as the states of the other robots in the team. To optimize this solution we develop a CSLAM back-end based on Consensus ADMM called MESA (Manifold, Edge-based, Separable ADMM). MESA is fully distributed to tolerate failures of individual robots, asynchronous to tolerate communication delays and outages, and general purpose to handle any CSLAM problem formulation. We demonstrate that MESA exhibits superior convergence rates and accuracy compare to existing state-of-the art CSLAM back-end optimizers.
\end{abstract}

\input{sections/0-intro}

\input{sections/1-related_work}
\input{sections/2-preliminaries}
\input{sections/3-methodology}

\input{sections/4-experiments}

\input{sections/5-conclusion}

\balance
\bibliographystyle{IEEEtran} 
\bibliography{refs}

\end{document}

%% file: helpers.tex
\usepackage{amsfonts}
\usepackage{amsthm}

\newcommand{\mat}[1]{\begin{bmatrix}#1\end{bmatrix}}

\DeclareMathOperator*{\argmax}{arg\,max}
\DeclareMathOperator*{\argmin}{arg\,min}

\newcommand{\logmap}[1]{\mathrm{Log}\left(#1\right)}

\newcommand{\norm}[1]{\left\|#1\right\|}

\newcommand{\inner}[2]{\left\langle #1, #2 \right\rangle}

\newcommand{\Cc}{\mathcal{C}}

\newcommand{\Ec}{\mathcal{E}}

\newcommand{\Lc}{\mathcal{L}}

\newcommand{\Rc}{\mathcal{R}}
\newcommand{\Sc}{\mathcal{S}}

\newcommand{\Zc}{\mathcal{Z}}

\newcommand{\Rb}{\mathbb{R}}



%% file: sections/0-intro.tex
\vspace{-10pt}
\section{Introduction}\label{sec:intro}

Collaborative teams of autonomous robots are desired across numerous application domains. Teams can work more efficiently than individual robots making them useful for applications like search and rescue~\cite{multirobot_sar_drew_2022} and scientific exploration~\cite{cadre_mission_website}. Moreover, heterogeneous teams of robots can use individual specializations to perform tasks beyond the capabilities of a single platform~\cite{arches_dlr_2020}. A key capability required for the effective deployment of multi-robot teams is Collaborative Simultaneous Localization and Mapping (CSLAM)~\cite{slam_survey_leonard_2016}. For downstream tasks like planning, robots need both an estimate of their own state as well as an estimate of states for relevant members of the team. In this work we focus on the CSLAM "back-end" responsible for composing noisy measurements into a state estimate for the robotic team.

Multi-robot teams require this state estimate to be both \textit{accurate} (i.e. optimal for the available information) and \textit{consistent} (i.e. robots agree on a single solution). Furthermore, teams require a method for computing this estimate that is \textit{efficient} with respect to both runtime and communication overhead as well as \textit{resilient} to practical communication network behavior like delayed/dropped messages or communication outages between robots. Finally, robotic teams need a method that is \textit{general purpose} so that teams can make use of all available map representations and sensing modalities.

An attractive approach to the CSLAM back-end is distributed optimization as it can adapt to failures of individual robots. Recent work has investigated the general classes of distributed optimization algorithms and observed that Consensus Alternating Direction Method of Multipliers (C-ADMM) displays superior convergence rates to alternative distributed optimization approaches~\cite{distsurvey_halsted_2021}. However, C-ADMM is yet unexplored in the context of CSLAM.

\begin{figure}[t!]
    \centering
    \vspace{5pt}
    \includegraphics[width=0.95\linewidth]{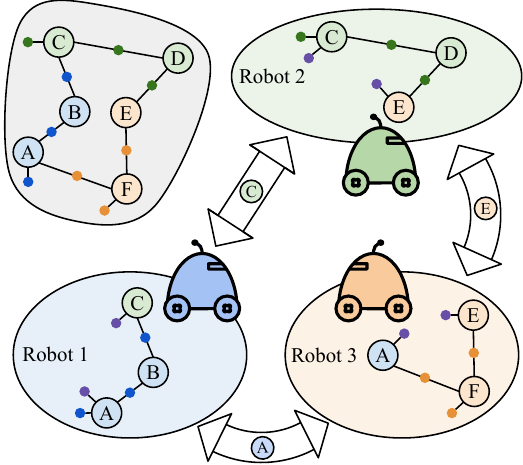}
    \caption{Example multi-robot factor graph (gray) and its corresponding distribution between agents (blue, green, and orange) in MESA. When asynchronously communicating (arrows), the robots send only their solutions for shared variables (A, C, E). These are used to construct "Biased Priors" (purple factors) incorporated into each robot's local graph to enforce consistency across the team.}
    \label{fig:mesa_high_level_example}
\end{figure}

\textbf{Contributions}: In this paper we first revisit C-ADMM and discuss extensions relevant to our proposed algorithm. We then propose a novel algorithm based on C-ADMM called \textbf{M}anifold, \textbf{E}dge-based, \textbf{S}eparable \textbf{A}DMM (MESA), an efficient, general purpose CSLAM back-end that permits asynchronous communication (see Fig.~\ref{fig:mesa_high_level_example}). We then explore variants of MESA derived from different manifold constraint formulations. We conclude by validating MESA under a variety of conditions seen in CSLAM problems and demonstrating that MESA achieves superior performance to existing distributed CSLAM back-end optimizers.

%% file: sections/1-related_work.tex
\section{Related Work}\label{sec:related-work}

ADMM was previously proposed to solve CSLAM problems by Choudhary et al. who proposed Multi-Block ADMM (MB-ADMM) to solve generic SLAM problems distributed on a cluster~\cite{choudhary_admm_2015}. MB-ADMM, however, is not resilient as it requires synchronized communication and unlike C-ADMM is proven to diverge for even some linear problems~\cite{mbadmm_bad_chen_2016}.

Other works have investigated alternate optimization methods for solving generic CSLAM problems. Cunningham et al. proposed Distributed Data Fusion Smoothing and Mapping (DDF-SAM) and DDF-SAM2. These algorithms share marginal information between robots being careful to avoid double counting~\cite{ddfsam_cunningham_2010, ddfsam2_cunningham_2013}. These approaches, however, fail to enforce equal linearization points between robots resulting in potentially inconsistent and inaccurate solutions. In another vein, Murai et al. proposed a Loopy Believe Propagation (LBP) based method~\cite{robot_web_murai_2022, loopy_bp_murphy_1999}. While distributed LBP demonstrates good performance, its convergence rate and consistency are unexplored in currently published work.

Another line of research has focused on distributed Pose Graph Optimization (PGO). PGO is a subset of CSLAM problems making these methods limited in their applicability. Choudhary et al. proposed the Distributed Gauss-Seidel (DGS) algorithm~\cite{dgs_choudhary_2017}. DGS solves the chordal relaxation of the PGO problem~\cite{chordal_init_carlone_2015} using distributed successive over relaxation. DGS often performs well, however, it requires synchronous communication and thus is not resilient to unreliable networks. Furthermore, it optimizes an approximation of the PGO objective, limiting its accuracy. Cristofalo et al. proposed another PGO method called GeoD which avoids approximation and optimizes the "geodesic" PGO formulation enabling better accuracy~\cite{geod_cristofalo_2020}. GeoD treats each pose as an independent agent and optimizes via distributed gradient descent making it inefficient and slow to converge.

Recent work on distributed PGO relaxes PGO to a Semi-Definite Program (SDP)~\cite{dc2pgo_tian_2021}. The proposed algorithm DC2-PGO and its asynchronous variant ASAPP solve CSLAM as a SDP with distributed Riemannian Gradient Descent (RGD). DC2-PGO and ASAPP are very accurate, can provide certificates of correctness, and ASAPP's asynchronous communication model makes it resilient to network issues~\cite{asapp_tian_2020}. However, like GeoD, by optimizing via distributed gradient descent DC2-PGO and ASAPP exhibit slow convergence.

As we can see, prior distributed SLAM methods struggle to achieve a sufficient solution. General purpose methods lack either consistency, accuracy, or resiliency. While some PGO methods achieve all of these goals they are slow to converge and limited in their applicability. The proposed algorithm, MESA, strives to achieve all of these goals.

%% file: sections/2-preliminaries.tex
\section{Consensus ADMM}\label{sec:cadmm}
We begin by reviewing Consensus ADMM~\cite{cadmm_mateos_2010}, a popular fully distributed optimization method for solving consensus optimization problems of the form in \eqref{eq:distributed_consensus_optimization_problem}.

\begin{equation}
\label{eq:distributed_consensus_optimization_problem}
\begin{aligned}
    \argmin_{\bar{x}\in\Rb^{|\Rc|n}} \quad & \sum_{i\in \Rc} f_i(x_i) \\
    \textrm{s.t.} \quad & x_i = x_j ~\forall~ (i,j) \in \Ec\\
\end{aligned}
\end{equation}

Where $\Rc$ is the set of all agents, $f_i$ is the local objective of agent $i$, $x_i \in  \Rb^n$ is the copy of decision variables held by agent $i$, and $\bar{x}\in \Rb^{|\Rc|n}$ is the concatenation of all agent's local decision variables. Agents communicate over an undirected network made up of the agents~$\Rc$ and communication links~$\Ec$. The constraints force neighboring agents to agree on a solution. By induction this forces all agents to converge to a single joint solution to the problem.

Applying standard ADMM to problem \eqref{eq:distributed_consensus_optimization_problem} produces an algorithm that requires centralized updates~\cite{boyd_distributed_2010}. To produce a fully distributed algorithm, C-ADMM augments problem \eqref{eq:distributed_consensus_optimization_problem} with additional variables. For each edge in the communication network $(i,j) \in \Ec$, C-ADMM introduces two new variables $z_{(i,j)}$ and $z_{(j,i)}$. The variable $z_{(i,j)}$ is held by agent $i$ and can be interpreted as agent $i$'s estimate of agent $j$'s local solution. We will refer to these variables as "edge variables" and use them rewrite \eqref{eq:distributed_consensus_optimization_problem} as \eqref{eq:consensus_admm_problem}.

\begin{equation}
\label{eq:consensus_admm_problem}
\begin{aligned}
    \argmin_{\bar{x}\in\Rb^{|\Rc|n},~\bar{z} \in \Zc} \quad & \sum_{i\in \Rc} f_i(x_i) \\
    \textrm{s.t.} 
        \quad & x_i = z_{(i,j)},~x_j = z_{(j,i)} ~\forall~ (i,j) \in \Ec
\end{aligned}
\end{equation}

Where $\bar{z}$ is the concatenation of all $z_{(i,j)}$ and $\Zc$ is the space of $\Rb^{2|\Ec|n}$ which has been further constrained such that $z_{(i,j)} = z_{(j,i)}$. This augmentation increases the number of constraints but does not change their meaning. C-ADMM solves \eqref{eq:consensus_admm_problem} according to the update process (\ref{eq:standard_admm_updates_x},~\ref{eq:standard_admm_updates_z},~\ref{eq:standard_admm_updates_dual},~\ref{eq:standard_admm_updates_beta}).

\begin{align}
    \label{eq:standard_admm_updates_x}
    & \bar{x}^{k+1} = \argmin_{\bar{x}\in\Rb^{|\Rc|n}}\Lc(\bar{x}, \bar{z}^k, \bar{\lambda}^k, \beta^k)\\
    \label{eq:standard_admm_updates_z}
    & \bar{z}^{k+1} = \argmin_{\bar{z} \in \Zc}\Lc(\bar{x}^{k+1}, \bar{z}, \bar{\lambda}^k, \beta^k)\\
    \label{eq:standard_admm_updates_dual}
    & \bar{\lambda}^{k+1} = \bar{\lambda}^{k} + \beta^k (D \bar{x}^{k+1} - \bar{z}^{k+1}) \\
    \label{eq:standard_admm_updates_beta}
    & \beta^{k+1} = \alpha \beta^k
\end{align}

Where $\bar{\lambda}$ is the concatenation of all dual variables (one corresponding to each constraint), $D\in \Rb^{(2|\Ec|n)\times(|\Rc|n)}$ maps each $x_i$ in $\bar{x}$ to all corresponding $z_{(i,j)}$ in $\bar{z}$, $\beta$ is the penalty term, $\alpha$ is a scaling factor hyper-parameter, $k$ is the iteration count, and $\Lc$ is the problem's Augmented Lagrangian \eqref{eq:standard_admm_lagrangian}.

\begin{multline}
\label{eq:standard_admm_lagrangian}
    \sum_{i\in \Rc} f_i(x_i) + \sum_{j \in \mathcal{N}_i} \inner{\lambda_{(i,j)}}{x_i - z_{(i,j)}} + \frac{\beta}{2}\norm{x_i - z_{(i,j)}}^2
\end{multline}

Where $\mathcal{N}_i$ are the neighbors of agent $i$. Solving these iterates can be fully distributed as \eqref{eq:standard_admm_updates_x} can be solved by each agent minimizing its local Augmented Lagrangian independently. After this minimization C-ADMM stipulates that all agents communicate their results to all neighbors to allow each agent to independently solve \eqref{eq:standard_admm_updates_z}, \eqref{eq:standard_admm_updates_dual}, and \eqref{eq:standard_admm_updates_beta}. When all $f_i$ are convex C-ADMM converges linearly~\cite{cadmm_lin_converge_shi_2014}.

Within existing literature there are two extensions to the base C-ADMM algorithm that will be relevant to MESA. Firstly, C-ADMM assumes that each agent's objective relies on all optimization parameters. However, there are a large class of problems where each agent's objective relies on only a subset of the variables. We refer to such problems as "separable" consensus optimization problems, a reference to Separable Optimization Variable ADMM (SOVA)~\cite{sova_shorinwa_2020}. SOVA proposed a simple but effective extension in which agents hold only the variables required by their local objective and we impose constraints on only the variables shared between agents. Separable problems take the form of \eqref{eq:sova_problem}.

\begin{equation}
\label{eq:sova_problem}
\begin{aligned}
    \argmin_{\{x_0 \in\Rb^{n_0},~...,~x_r \in\Rb^{n_r} \}} \quad & \sum_{i\in \Rc} f_i(x_i) \\
    \textrm{s.t.} \quad & A_{(i,j)} x_i = B_{(i,j)} x_j ~\forall~ (i,j) \in \Ec\\
\end{aligned}
\end{equation}

Where $A_{(i,j)}\in \Rb^{n_{(i,j)} \times n_{i}}$ and $B_{(i,j)}\in \Rb^{n_{(i,j)} \times n_{j}}$ map the variables shared between agents $i$ and $j$ into a common space.

Secondly, C-ADMM assumes an algorithmic structure in which, at each iteration, all agents communicate with all neighbors. However, C-ADMM has been proven to converge for significantly less restrictive communication models. Edge-based C-ADMM assumes that at each iteration of the algorithm only a subset of communication edges are active and each agent, therefore, communicates with only some of its neighbors~\cite{thesis_wei_2014}. Without loss of generality this model can be simplified to assume that at each iteration only two agents communicate with each other. Under this Edge-based communication model C-ADMM is proven to convergence with rate $O(1/k)$ where $k$ is the number of iterations~\cite{o1k_convergence_wei_2013}.

%% file: sections/3-methodology.tex
\section{Methodology}\label{sec:methodolody}

In this section we first define the CSLAM problem and demonstrate that it can be transformed into an instance of a Separable Consensus ADMM problem with on-manifold decision variables. We then propose a novel algorithm MESA for solving general CSLAM problems with asynchronous communication. Finally, we derive variants of the algorithm driven by different approaches to modeling constraints between the manifold decision variables found in CSLAM.

We focus on solving generalized CSLAM problems, which can be defined as Maximum-A-Posterori (MAP) inference.

\begin{equation} \label{eq:cslam_as_map}
    \Theta_{MAP} = \argmax_{\Theta\in\Omega} P(\Theta | M)
\end{equation}

Where $\Theta$ are all the variables of interest, $\Omega$ is the product manifold formed by the manifold of each element within $\Theta$, and $M$ is the set of measurements. When we assume our measurements are affected by Gaussian noise (i.e. $m \sim \mathcal{N}\left(\mu_m, \Sigma_m\right)$), problems of this form can be solved via nonlinear least squares optimization~\cite{factor_graphs_for_robot_perception}.

\begin{equation} \label{eq:cslam_as_nls}
    \Theta_{MAP} = \argmin_{\Theta \in \Omega} \sum_{m\in M} \norm{h_m(\Theta) - m}_{\Sigma_m}^2
\end{equation}

Where $h_m$ is the measurement prediction function that computes the expected $m$ from an estimate of the state.

\subsection{Manifold, Edge-based, Separable ADMM (MESA)}

In the case of a multi-robot team $\Rc$, the variables $\Theta$ and measurements $M$ are distributed across robots. Let $M_i$ denote the measurements made by robot $i$ and $\Theta_i$ denote the variables that are observed by the measurements in $M_i$. Multiple robots may observe the same variable (e.g. two robots observe the same landmark). Therefore, while $M_i$ are disjoint subsets of $M$, each $\Theta_i$ is a non-disjoint subset of $\Theta$. Let $\Sc_{(i,j)} \triangleq \Theta_i \cap \Theta_j$ represent the variables shared between robots $i$ and $j$ where for $\theta_{s} \in \Sc_{(i,j)}$ we denote $\theta_{s_i}$ as the copy of that variable owned by robot $i$. We can therefore re-write \eqref{eq:cslam_as_nls} to reflect this distribution of information.

\begin{equation} \label{eq:cslam_as_distributed_opt}
\begin{aligned}
    \Theta_{MAP} = \argmin_{\Theta_i \in \Omega_i \forall i\in\Rc}\quad& 
        \sum_{i\in \Rc} \sum_{m\in M_i} \norm{h_m(\Theta_i) - m}_{\Sigma_m}^2 \\
        \textrm{s.t.} \quad & q_{s}(\theta_{s_i}, \theta_{s_j}) = 0 \\
                            &~\forall~ \theta_s \in \Sc_{(i,j)} ~\forall~(i,j) \in \Ec\\
\end{aligned}
\end{equation}

Where $q_{s}$ is a function that compares the equality appropriately for the manifold to which $\theta_{s}$ belongs and returns $0 \in \Rb^d$ if and only if $\theta_{s_i}$ and $\theta_{s_j}$ are equal. The generic function $q_{s}$ is used as there exists potentially many ways to compare equality of on-manifold objects. Concrete implementations of $q_s$ will be explored in detail in Sec.~\ref{sec:methodology:constraint_functions}.

From \eqref{eq:cslam_as_distributed_opt} we can see that the general CSLAM problem is an instance of a separable optimization problem \eqref{eq:sova_problem} as each robot's cost function affects only a subset of the global variables and robots share sparse sets of variables. However, unlike \eqref{eq:sova_problem} and all standard ADMM formulations, CSLAM problems typically include on-manifold decision variables which we handle with generic constraint functions $q_s$.

To make the CSLAM problem fully distributed we apply the same method as C-ADMM and augment the problem with edge variables. Specifically, for every shared variable $\theta_s$ shared along edge $(i,j)$ we introduce edge variables $z_{(i,j)_s}$ and $z_{(j,i)_s}$. The addition of edge variables allows us to re-write the constraints of our problem as follows in \eqref{eq:cslam_as_ebadmm}.

\begin{equation} \label{eq:cslam_as_ebadmm}
\begin{aligned}
    \Theta_{MAP} = \argmin_{\Theta_i \in \Omega_i \forall i\in\Rc,~Z \in \Zc}\quad& 
        \sum_{i\in \Rc} \sum_{m\in M_i} \norm{h(\Theta_i) - m}_{\Sigma_m}^2 \\
        \textrm{s.t.} \quad & q_{s}(\theta_{s_i}, z_{(i,j)_s}) = 0 \\
                            & q_{s}(\theta_{s_j}, z_{(j,i)_s}) = 0 \\
                            &~\forall~ \theta_s \in \Sc_{(i,j)} ~\forall~(i,j) \in \Ec\\
\end{aligned}
\end{equation}

Where $Z$ is the set of all $z_{(i,j)_s}$ and $\Zc$ is the appropriate product manifold further constrained such that $z_{(i,j)_s} = z_{(j,i)_s}$. Next we use 
(\ref{eq:standard_admm_updates_x},~\ref{eq:standard_admm_updates_z},~\ref{eq:standard_admm_updates_dual},~\ref{eq:standard_admm_updates_beta},~\ref{eq:standard_admm_lagrangian}) to derive on-manifold, separable, C-ADMM iterates required to solve \eqref{eq:cslam_as_ebadmm}.

\begin{align}
    & \label{eq:mesa_updates_x}
    \begin{aligned}
    \Theta^{k+1}_{i} =& \argmin_{\Theta_i \in \Omega_i} \quad 
          \sum_{m\in M_i} \norm{h_m (\Theta^k_i) - m}_{\Sigma_m}^2 \\
          &+ \sum_{j \in \mathcal{N}_i} \sum_{s\in \Sc_{(i, j)}} \inner{\lambda^k_{(i,j)_s}}{q_{s}\left(\theta_{s_i}, z^{k}_{(i,j)_s}\right)} \\
          &+ \sum_{j \in \mathcal{N}_i} \sum_{s\in \Sc_{(i, j)}} \frac{\beta^k}{2} \norm{q_{s}\left(\theta_{s_i}, z^{k}_{(i,j)_s}\right)}^2\\
    \end{aligned}\\
    & \label{eq:mesa_updates_z}
    \begin{aligned}
         {z^{k+1}_{(i, j)_s}}& = 
          \argmin_{z_s \in \Zc_s}\\ 
          &\inner{\lambda^k_{(i,j)_s}}{q_{s}\left(\theta^{k+1}_{s_i}, z_s\right)} + \frac{\beta^k}{2} \norm{q_{s}\left(\theta^{k+1}_{s_i}, z_s\right)}^2 \\
         +&\inner{\lambda^k_{(j,i)_s}}{q_{s}\left(\theta^{k+1}_{s_j}, z_s\right)} + \frac{\beta^k}{2} \norm{q_{s}\left(\theta^{k+1}_{s_j}, z_s\right)}^2 \\
    \end{aligned}\\
    & \label{eq:mesa_updates_dual}
    \lambda_{(i,j)_s}^{k+1} = \lambda_{(i,j)_s}^{k} + \beta^k \left(  q_{s}\left(\theta^{k+1}_{s_i}, z^{k+1}_{(i,j)_s}\right) \right) \\
    \label{eq:mesa_updates_beta}
    & \beta^{k+1} = \alpha \beta^k
\end{align}

Where we have introduced a dual variable $\lambda_{(i,j)_s}$ for each constraint and written the iterates for individual variables where they can be solved for independently.

Using these iterates with a C-ADMM algorithm, however, is insufficient for typical CSLAM problems. Multi-robot teams operating in the field cannot assume reliable network infrastructure. Therefore, robots may not be able to synchronously communicate with all neighbors as is required by C-ADMM. To make our algorithm resilient to these conditions we combine these iterates  with an Edge-based communication model. Within the Edge-based model, communication dropouts and message delay/loss are modeled simply as iterations in which the affected robots do not communicate. The resulting asynchronous algorithm can tolerate expected network conditions. To fully address asynchronous communication we further modify the iterates adding unique penalty terms $\beta_{(i,j)}, \beta_{(j,i)}$ for each edge in $\Ec$.

The edge-based communication model combined with our on-manifold, separable, C-ADMM iterates produces MESA (Alg.~\ref{alg:mesa}) which is also outlined in Fig.~\ref{fig:mesa_high_level_example}.
\begin{algorithm}
\footnotesize 
\captionsetup{font=footnotesize} 
\caption{Manifold, Edge-based, Separable, ADMM (MESA) }\label{alg:mesa}
    \begin{algorithmic}[1]
    \State \textbf{In:} Robots $\Rc$, communication links $\Ec$, local estimates $\left\{\Theta^0_0, \Theta^0_1, ..., \Theta^0_r\right\}$
    \State \textbf{Out:} Final Variable Estimates $\left\{\Theta^{final}_0, \Theta^{final}_1, ..., \Theta^{final}_r\right\}$
    \State $\lambda_{(i,j)_s}, \lambda_{(j,i)_s} \gets \mathbf{0}~\forall~ s \in \Sc_{(i,j)} ~\forall~ (i,j) \in \Ec$
    \State $z_{(i,j)_s} \gets \theta^0_{s_i} , z_{(j,i)_s} \gets \theta^0_{s_j} ~\forall~ s \in \Sc_{(i,j)} ~\forall~ (i,j) \in \Ec$
    \While{Not Converged}
        \If{Communication available between robot $i$ and robot $j$}
            \State In parallel update $\Theta_i$ and $\Theta_j$ with \eqref{eq:mesa_updates_x}
            \State Between $i$ and $j$ communicate $\theta_{s_i}, \theta_{s_j} ~\forall~\theta_s \in \Sc_{(i,j)}$
            \State In parallel update $z_{(i,j)_s}, z_{(j,i)_s} ~\forall~\theta_s \in \Sc_{(i,j)}$ with \eqref{eq:mesa_updates_z}
            \State In parallel update $\lambda_{(i,j)_s}, \lambda_{(j,i)_s} ~\forall~\theta_s \in \Sc_{(i,j)}$ with \eqref{eq:mesa_updates_otherdual}
            \State In parallel update $\beta_{(i,j)}^{k+1} = \alpha \beta_{(i,j)}^k$ and $\beta_{(j,i)}^{k+1} = \alpha \beta_{(j,i)}^k$
        \EndIf
    \EndWhile
    \end{algorithmic}
\end{algorithm}

\begin{remark}[MESA Initialization]\label{remark:mesa_initialization}
Without loss of generality we define $\lambda^0_{(i,j)_s} = \mathbf{0}$ where $\mathbf{0}$ is of proper dimension. Further, we assume $z_{(i,j)_s} = \theta_{s_i}$ until the first communication between robots $i$ and $j$.
\end{remark}

\begin{remark}[MESA Theoretical Guarantees]\label{remark:theoretical_guarentees}
Due to the non-convex objective and non-convex variable space found in CSLAM, MESA does not share the convergence guarantees of C-ADMM, SOVA, or Edge-based C-ADMM. Recent work has proven that, under certain assumptions,  non-convex C-ADMM will converge~\cite{admm_nonconvex_converge_hong_2016}. However, this proof does not addressed on-manifold decision variables. More importantly, for practical CSLAM it is unclear whether the problems will meet the required assumptions. Therefore, we do not pursue a convergence proof for MESA. Instead we opt to demonstrate convergence  of MESA empirically (see Sec.~\ref{sec:experiments}).
\end{remark}

\subsection{MESA Implementation}
Implementation of Alg.~\ref{alg:mesa} requires solving the optimization problems \eqref{eq:mesa_updates_x}  and \eqref{eq:mesa_updates_z}.  As originally demonstrated by Choudhary et al.~\cite{choudhary_admm_2015} the Augmented Lagrangian in \eqref{eq:mesa_updates_x} can be represented as a factor-graph by re-writing the dual and penalty terms as "Biased Priors". Specifically, the identity that $\argmin_a \inner{b}{a} +(\beta/2)\norm{a}^2 = \argmin_a (\beta/2)\norm{a + b/\beta}^2$ when $b$ is constant is used to re-write the terms as \eqref{eq:biased_prior}.

\begin{equation}
\label{eq:biased_prior}
\frac{\beta_{(i,j)}}{2} \norm{q_{s}\left(\theta_{s}, z_{(i,j)_s}\right) + \frac{\lambda_{(i,j)_s}}{\beta_{(i,j)}}}^2 
\end{equation}

This allows us to compute \eqref{eq:mesa_updates_x} using existing sparse nonlinear least squares libraries like \texttt{gtsam}~\cite{dellaert_gtsam_tech_report_2012}.

The edge variable update \eqref{eq:mesa_updates_z} could be implemented similarly and each $z_{(i,j)_s}$ independently solved via a nonlinear optimization. However, to improve efficiency, we would prefer that \eqref{eq:mesa_updates_z} be solved in closed form. Depending on the selection of constraint function and edge variable space, a closed form solution may exist or may be approximated. 

Computing updates to dual variables is straightforward according to \eqref{eq:mesa_updates_dual}. In this, dual variables are updated based on the error between a $\theta_{s_i}$ and $z_{(i,j)_s}$. However, this error underestimates the magnitude of the constraint satisfaction gap (i.e. the error between $\theta_{s_i}$ and $\theta_{s_j}$). We observe that updating dual variables according to \eqref{eq:mesa_updates_otherdual} is more reflective the proper magnitude and improves convergence speed.
\begin{equation}
    \lambda_{(i,j)_s}^{k+1} = \lambda_{(i,j)_s}^{k} + \beta^k_{(i,j)} \left(  q_{s}\left(\theta^{k+1}_{s_i}, \theta^{k+1}_{s_j}\right) \right)
    \label{eq:mesa_updates_otherdual}
\end{equation}

\subsection{Manifold Constraint Functions} \label{sec:methodology:constraint_functions}
To compute \eqref{eq:mesa_updates_x}, \eqref{eq:mesa_updates_z}, and \eqref{eq:mesa_updates_otherdual} we must also concretely define the constraint function used for each shared variable. The selection of constraint function $q_s$ depends on the type of our shared variables $\theta_s$ and the type selected for edge variables $z_{(i,j)_s}$. In the existing C-ADMM literature (see Sec.~\ref{sec:cadmm}) all algorithms assume vector-valued decision variables and thus that all constraints are linear ($q(a,b) \triangleq a-b$) which permits a closed form solution to \eqref{eq:mesa_updates_z} of $z_{(i,j)_s}^{k+1} = \frac{1}{2}(\theta^{k+1}_{s_i} + \theta^{k+1}_{s_j})$. For linear variables in CLSAM (e.g. landmarks) this approach should be taken. 

However, in CSLAM we are typically also optimizing over robot poses that are contained on the $\mathrm{SE}(N)$ manifold. For such on-manifold variables we identify four unique constraint functions (Geodesic, Approximate-Geodesic, Split, and Chordal). For each we derive the corresponding closed form solution to \eqref{eq:mesa_updates_z} or an approximate solution if no closed form solution exists. A summary of constraint functions and shared variable updates can be found in Table~\ref{tab:constraint_functions}.

\begin{table}[ht]
\caption{Constraint functions for $\mathrm{SE}(N)$ objects and their corresponding closed form solutions to \eqref{eq:mesa_updates_z}. Where SPLIT interpolates the translation component linearly and the rotation component spherically~\cite{interpolation_survey_haarbach_2018}, Vec returns a vector of the objects non-constant matrix elements~\cite{madamm_kovnatsky_2016}, and $p$ is the dimension of the tangent space for SE$(N)$. Manifold object notation is derived from the work of Solà et al. \cite{microlie_sola_2021}. The geodesic $z$ update is approximated by the case where $\lambda_{(i,j)_s} = \lambda_{(j,i)_s} = \mathbf{0}$.}
\centering
\footnotesize
\begin{tabular}{|Sl|Sc|Sc|Sc|}
\hline
Function                & $z\in$            & $q(\theta, z)$                                & $z^{k+1}_{(i,j)_s}$ \\ \hline\hline
Geodesic            & $\mathrm{SE}(N)$  & $\logmap{z^{-1} \circ \theta}$                & $\mathrm{SPLIT}\left(\theta_{s_i}, \theta_{s_j}, 0.5\right)$ \\  \hline
Apx.-Geo.     & $\Rb^{p}$         & $\logmap{\theta} - z$                         & $\frac{1}{2}\left(\logmap{\theta_{s_i}} + \logmap{\theta_{s_j}}\right)$  \\ \hline
Split               & $\mathrm{SE}(N)$  & $\mat{\logmap{R_z^{-1}R_\theta} \\ t_\theta - t_z}$    & $\mathrm{SPLIT}\left(\theta_{s_i}, \theta_{s_j}, 0.5\right)$ \\ \hline
Chordal             & $\Rb^{N^2+N}$   & $\mathrm{Vec}(\theta) - z$                    & $\frac{1}{2}\left( \mathrm{Vec}\left(\theta_{s_i}\right) + \mathrm{Vec}\left(\theta_{s_j}\right)\right)$ \\ \hline
\end{tabular}
\label{tab:constraint_functions}
\end{table}

Each constraint function produces a slight variant of the MESA algorithm. We identify each variant by the name of the constraint function used (i.e. Geodesic MESA is the MESA algorithm that uses the Geodesic constraint).

\begin{remark}[Hyper-Parameters]\label{remark:mesa_parameters}
All MESA variants use $\beta_{(i,j)}^0 = 200$ and $\alpha=1$ with the exception of Split MESA that uses $\alpha = 1.2$. Note that these parameters are modified for some experiments.
\end{remark}

%% file: sections/4-experiments.tex
\section{Experiments}\label{sec:experiments}
In this section we evaluate the performance of MESA to solve representative CSLAM problems. We first compare the four variants of the algorithm induced by choice of constraint function. With the best performing variants we then explore the effect of problem parameters (length, initialization quality, number of robots, and types of measurements) to validate its performance across different CSLAM scenarios. Finally, we compare MESA to state of the art works and demonstrate its superior convergence compared to existing methods.

\subsection{Experiment Setup}

\subsubsection{Dataset Generation}
\begin{figure}[ht]
    \centering
    \begin{subfigure}{0.4\linewidth}
      \centering
      \includegraphics[width=0.9\linewidth, trim={1.5cm 1.5cm 1.5cm 1.5cm}, clip]{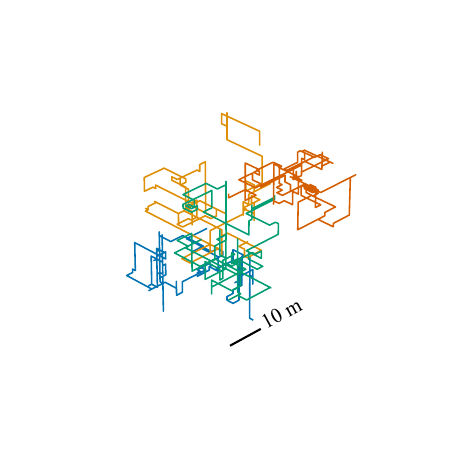}
      \caption{Example 3D Dataset}
      \label{fig:example_trajectories:length}
    \end{subfigure}%
    \begin{subfigure}{0.4\linewidth}
      \centering
      \includegraphics[width=0.9\linewidth]{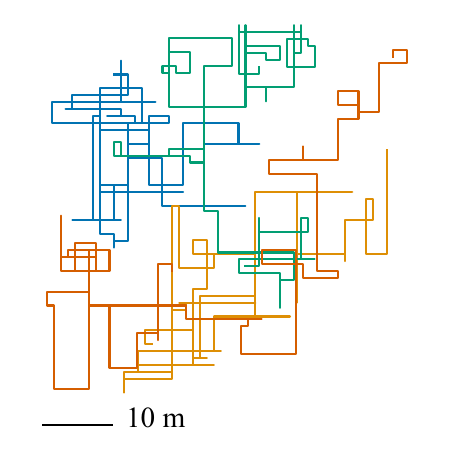}
      \caption{Example 2D Dataset}
      \label{fig:example_trajectories:range}
    \end{subfigure}%
    \vspace{-0.2cm}
    \caption{Ground-truth from example synthetic datasets. Colors represent different robots.}
    \label{fig:example_synthetic_dataset}
\end{figure}

To explore different CSLAM conditions we generate datasets by sampling odometry measurements from a categorical distribution over forward motion and a $\pm90^\circ$ rotation around each axis. At each time-step robots add intra-robot loop closures with probability $p=0.4$ if the current pose is nearby a previous pose. Every $k$ steps we search for inter-robot loop closures. Unless modified, each experiment consists of 10 datasets generated with 4 robots traversing a 400 pose long trajectory. All measurements are relative poses subject to a noise model of $(\sigma_r = 1^\circ, \sigma_t = 0.05m)$. With this general framework we can generate a variety of 2D and 3D datasets (e.g. Fig~\ref{fig:example_synthetic_dataset}).

\subsubsection{Metric}
Single robot optimization tasks are often evaluated using the cost function residual~\eqref{eq:cslam_as_nls}. In multi-robot problems, however, we cannot directly apply this metric. For each \textit{shared variable}, there exists a local copy on each robot that observed the variable. We want a \textit{consistent} solution such that all local copies are equal. However, practically local copies will differ slightly and it is ambiguous which copy to use to compute the residual. Moreover, in cases where local copies differ we want a metric that reflects the inconsistency. To meet these goals we introduce the \textit{Mean Residual} ($r^2_{mean}$). Mean residual is an extension of the SLAM cost function in which, the cost contribution for any factor that affects shared variables is averaged over the cost from all possible combinations of solutions. If all shared variables agree exactly, $r^2_{mean}$ is the same as the SLAM cost function and represents the accuracy of the optimized solution. If shared variables disagree, $r^2_{mean}$ still captures this accuracy but also monotonically increases with any shared variable disagreement. Concretely we define $r^2_{mean}$ as:

\begin{equation}
    r^2_{mean} = \sum_{m\in M} \frac{1}{|\Cc|}\sum_{(i,j, ...)\in \Cc} \norm{h_m\left(\theta_{a_i}, \theta_{b_j}, ...\right) - m}_{\Sigma_m}^2    
\end{equation}

Where, for a factor on variables $\{\theta_a, \theta_b ...\}$, $\Cc$ represents the set of combinations of local solutions to these variables. For example if a factor affects variables $a$ and $b$ where $a$ is shared between robots $\{i,j\}$ and $b$ is shared between robots $\{k,l\}$ then $\Cc = \{ (i, k), (i, l), (j, k), (j, l)\}$. If $a$ and $b$ are not shared and owned by robot $i$ then $\Cc$ is simply $\{(i,i)\}$.

\subsection{MESA Variant Exploration}
In our first experiment we explore the performance of different variants of MESA. This experiment is run on 20 random 3D datasets and a summary of $r^2_{mean}$ achieved by each variant are represented by box-plots.

\begin{figure}[ht]
    \centering
    \includegraphics[width=0.9\linewidth]{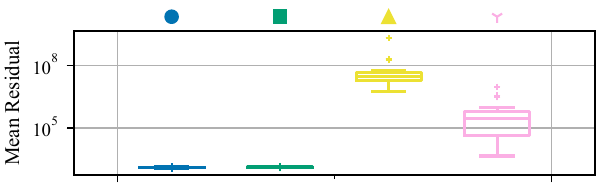}
    \caption{Comparison of MESA variants derived from Geodesic (\SymMesaGeo), Split (\SymMesaSplit), Approximate-Geodesic (\SymMesaOneOrd) and Chordal (\SymMesaManif) constraints on 20 synthetic 3D pose graph datasets. Split and Geodesic constraints significantly outperform the alternatives.}
    \label{fig:mesa_variant_experiment}
\end{figure}

As can be seen in Fig.~\ref{fig:mesa_variant_experiment} the selection of constraint function has a significant affect on the performance of the algorithm. We can see that the Geodesic and Split constraints significantly outperform the Chordal and Approximate-Geodesic constraints. Throughout the rest of this paper we will present only results from the Geodesic and Split variants.

\subsection{MESA Generalization}
Next we look to validate that MESA generalizes to different CSLAM scenarios. For each scenario we present results in comparison to two baselines: a Centralized approach in which all measurements are aggregated and solved using Levenberg-Marquardt and an Independent approach in which robots ignore all inter-robot measurements and solve their local factor graph independently. Specifically, we explore:

\begin{figure*}[t]
    \centering
    \begin{subfigure}{0.5\textwidth}
      \centering
      \begin{overpic}[width=0.8\linewidth]
      {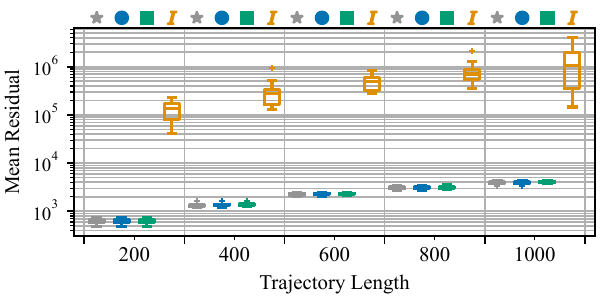}
        \put (-2,44) {\textbf{(a)}}
      \end{overpic}
      \phantomsubcaption
      \label{fig:dataset_variation_experiment:length}
    \end{subfigure}%
    \begin{subfigure}{0.5\textwidth}
      \centering
      \begin{overpic}[width=0.8\linewidth]
      {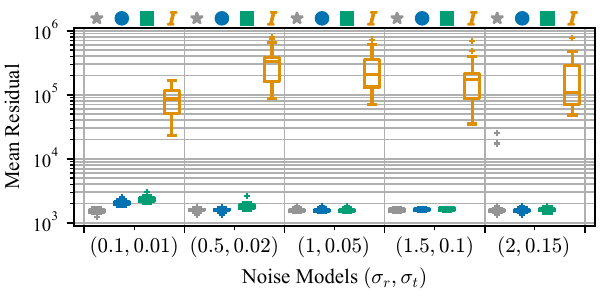}
        \put (-2,44) {\textbf{(b)}}
      \end{overpic}
      \phantomsubcaption
      \label{fig:dataset_variation_experiment:initialization}
    \end{subfigure}%
    \hfill
    \begin{subfigure}{0.5\textwidth}
      \centering
      \begin{overpic}[width=0.8\linewidth]
      {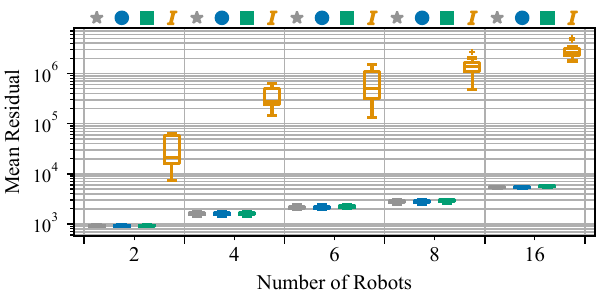}
        \put (-2,44) {\textbf{(c)}}
      \end{overpic}
      \phantomsubcaption
      \label{fig:dataset_variation_experiment:scale}
    \end{subfigure}%
    \begin{subfigure}{0.5\textwidth}
      \centering
      \begin{overpic}[width=0.8\linewidth]
      {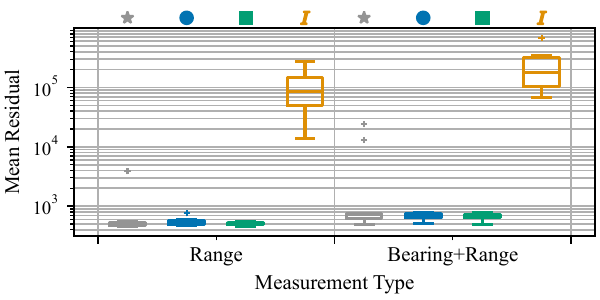}
        \put (-2,44) {\textbf{(d)}}
      \end{overpic}
      \phantomsubcaption
      \label{fig:dataset_variation_experiment:measurement_type}
    \end{subfigure}%
    \vspace{-0.2cm}
    \caption{Performance of Geodesic MESA (\SymMesaGeo) and Split MESA (\SymMesaSplit) compared to Centralized (\SymCent) and Independent (\SymIndep) baselines under a variety of conditions. The MESA variants consistently converge to the centralized solution indicating they are accurate and generalize across different conditions found in CSLAM problems.}
    \label{fig:dataset_variation_experiment}
    \vspace{-0.8cm}
\end{figure*}

\subsubsection{Trajectory Length (Fig.~\ref{fig:dataset_variation_experiment:length})}
First we evaluate how MESA handles different sizes of 3D factor graphs and vary trajectory length from 200 to 1000 poses for all robots.

\subsubsection{Problem Initialization (Fig.~\ref{fig:dataset_variation_experiment:initialization})}
Next we evaluate how initialization quality affects MESA. The quality of initialization from odometry is proportional to the magnitude of the noise model and thus we vary the measurement noise used to generate each dataset. This trial is run on 20 3D datasets.

\subsubsection{Problem Scale (Fig.~\ref{fig:dataset_variation_experiment:scale})}
We also inspect how MESA scales with the size of the robot team. To do so we vary the number of robots in each 3D dataset from 2 to 16.

\subsubsection{Measurement Models (Fig.~\ref{fig:dataset_variation_experiment:measurement_type})}
In this sub-experiment we evaluate the ability of MESA to handle generalized 2D CSLAM problems. We explore two types of inter-robot loop-closure measurements (range and bearing+range). These measurements are particularly compelling as they can be made locally and do not require any communication of raw sensor data to support a distributed front-end. For this scenario both MESA variants use $\beta_{(i,j)}^0=2$ and $\alpha=1.05$.

As can be seen in Fig.~\ref{fig:dataset_variation_experiment}, both the Geodesic and Split MESA variants generalize across different conditions found in CSLAM. In all scenarios the algorithms converge toward the centralized solution indicating that the algorithms are both accurate and reliable. Moreover, this experiment validates that collaboration between robots can significantly improve localization performance over independent operation.

\subsection{Prior Work Comparison}
In our final experiment we evaluate the performance of MESA relative to existing prior works namely DDF-SAM2~\cite{ddfsam2_cunningham_2013}, MB-ADMM~\cite{choudhary_admm_2015}, DGS~\cite{dgs_choudhary_2017}, and ASAPP~\cite{asapp_tian_2020}. Since ASAPP and DGS are distributed PGO algorithms we limit our experiments in this section to pose graph CSLAM. 

\begin{remark}[Prior Work Implementation Details]
DDF-SAM2 is implemented by the authors.  MB-ADMM is provided by the original authors\footnote{\scriptsize\url{https://github.com/itzsid/admm-slam}} and uses parameters from~\cite{choudhary_admm_2015}. DGS is provided by the original authors\footnote{\scriptsize\url{https://github.com/CogRob/distributed-mapper}} and uses default parameters. ASAPP is provided by the original authors\footnote{\scriptsize\url{https://github.com/mit-acl/dpgo}} and we use $100$ RGD steps per iteration with step size $0.001$. A small step size was used as larger step sizes sometimes caused divergence and we ran too many trials to permit tuning the step size for each dataset as done in~\cite{asapp_tian_2020}. All methods were limited to a maximum of $500 * |\Ec| * |\Rc|$ communications.
\end{remark}

\subsubsection{Benchmark Datasets}
We first compare the performance of these methods on partitioned versions of benchmark SLAM datasets. Each dataset is partitioned into 5 sub-graphs using METIS partitioning~\cite{choudhary_admm_2015}. Table~\ref{tab:benchmark_exp_results} compares the achieved $r^2_{mean}$. This sub-experiment omits Independent and DDF-SAM2 results as not all partitions have priors and these methods will be under-determined. In this sub-experiment both MESA variants use $\beta_{(i,j)}^0=2$ and $\alpha=1.05$.

\begin{table}[ht]
\centering
\caption{Mean residuals achieved by MESA and prior works on benchmark datasets: Sphere2500~\cite{isam_kaess_2007}, Parking Garage~\cite{chordal_init_carlone_2015}, and Torus~\cite{chordal_init_carlone_2015}.}
\label{tab:benchmark_exp_results}
\begin{tabular}{|Sl||Sc|Sc|Sc|}
\hline
Dataset                             & Sphere2500    & Parking Garage    & Torus \\ \hline\hline
Centralized                         & 675.7         & 0.634             & 29981 \\ \hline\hline
MB-ADMM~\cite{choudhary_admm_2015}  & 1.7e+22       & 0.638             & 2.5e+14 \\ \hline
DGS~\cite{dgs_choudhary_2017}       & 1204.9        & 0.650             & 12254 \\ \hline
ASAPP~\cite{asapp_tian_2020}        & 940.2         & 0.651             & \textbf{12196} \\ \hline
Geodesic MESA                       & \textbf{676.2}& \textbf{0.636}    & 30002 \\ \hline
Split MESA                          & 1054.7        & 0.645             & 30014 \\ \hline
\end{tabular}
\end{table}
\vspace{4pt}
Table~\ref{tab:benchmark_exp_results} illustrates that MESA is able to converge very closely to the centralized solution across all datasets. It also highlights that, like the centralized solution, MESA is a local solver and can fall into local optima like on the Torus dataset.

\subsubsection{Empirical Convergence}
We also explore the performance of each algorithm with respect to its convergence speed. As each algorithm's implementation differs, we propose a normalized measure of runtime -- the number of communications. We define one communication as when a pair of robots pass bidirectionally any amount of information. For example an iteration of Alg.~\ref{alg:mesa} results in one communication.

Communication is expected to be the largest bottleneck in the CSLAM pipeline and, for all algorithms, the number of communications is proportional to both computational cost and communication burden. Therefore, the number of communications provides a holistic metric to represent the complexity of the algorithm. To account for varying convergence criteria, we run all algorithms with a tight stopping condition and report results ($r^2_{mean}$ and communications) when the algorithm reaches within $1\%$ of its final $r^2_{mean}$. To reduce variance, this experiment is run on 200 3D datasets.

\begin{figure}[ht]
    \centering
    \includegraphics[width=0.85\linewidth]{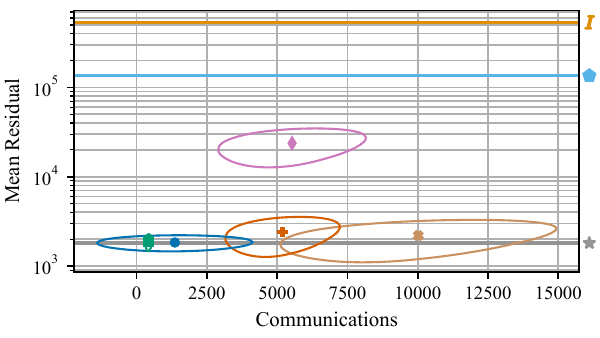}
    \caption{Accuracy vs. Communications of Geodesic MESA (\SymMesaGeo) and Split MESA (\SymMesaSplit) compared to prior works [DGS (\SymDgs), ASAPP (\SymAsapp), MB-ADMM (\SymMbAdmm), DDF-SAM2 (\SymDdf)] and baselines [Centralized (\SymCent), Independent (\SymIndep)] on 200 synthetic 3D pose graph datasets. Ellipses depict $3\sigma$ uncertainty bounds. Methods that require only one round of communication are shown as horizontal lines. MESA outperforms prior works both with respect to accuracy and convergence speed.
    }
    \label{fig:convergence_rate_experiment}
    \vspace{-0.15cm}
\end{figure}

As we can see in Fig.~\ref{fig:convergence_rate_experiment} both MESA variants exhibit significantly faster convergence than the prior works while providing superior accuracy. Though difficult to see in Fig.~\ref{fig:convergence_rate_experiment}, Split MESA converges more quickly than Geodesic MESA at the cost of reduced accuracy. ASAPP, despite being certifiably correct, exhibits worse performance than MESA as it often reached the communication limit before fully converging. This experiment also shows that, as expected, DGS provides less accurate results due to its chordal approximation of the PGO problem, MB-ADMM struggles to converge, and DDF-SAM2's performance lags other methods as it does not produce consistent solutions.

%% file: sections/5-conclusion.tex
\section{Conclusion and Future Work}\label{sec:discussion}
In this work we presented MESA, a distributed algorithm for solving general CSLAM problems. MESA exhibits a superior convergence rate and final accuracy compare to prior works and accomplishes this for generalized CSLAM problems while permitting asynchronous communication. 

While MESA demonstrates excellent performance across experiments, the quality of its convergence can be dependent on hyper-parameters $\beta^0$ and $\alpha$. Determining these parameters automatically would be a good extension to MESA. Additionally, like most prior works, MESA is a batch solver. Therefore, despite its fast convergence it still requires many rounds of communication to compute a solution making it difficult to apply to real-time applications. Future work should include extending MESA to operate incrementally.

%% file: root.bbl
\begin{thebibliography}{10}
\providecommand{\url}[1]{#1}
\csname url@samestyle\endcsname
\providecommand{\newblock}{\relax}
\providecommand{\bibinfo}[2]{#2}
\providecommand{\BIBentrySTDinterwordspacing}{\spaceskip=0pt\relax}
\providecommand{\BIBentryALTinterwordstretchfactor}{4}
\providecommand{\BIBentryALTinterwordspacing}{\spaceskip=\fontdimen2\font plus
\BIBentryALTinterwordstretchfactor\fontdimen3\font minus
  \fontdimen4\font\relax}
\providecommand{\BIBforeignlanguage}[2]{{%
\expandafter\ifx\csname l@#1\endcsname\relax
\typeout{** WARNING: IEEEtran.bst: No hyphenation pattern has been}%
\typeout{** loaded for the language `#1'. Using the pattern for}%
\typeout{** the default language instead.}%
\else
\language=\csname l@#1\endcsname
\fi
#2}}
\providecommand{\BIBdecl}{\relax}
\BIBdecl

\bibitem{multirobot_sar_drew_2022}
D.~Drew, ``Multi-agent systems for search and rescue applications,''
  \emph{Current Robotics Reports}, vol.~2, pp. 189--200, 2021.

\bibitem{cadre_mission_website}
{NASA Jet Propulsion Labratory}, ``{Cooperative Autonomous Distributed Robotic
  Exploration (CADRE)},'' \url{https://www.jpl.nasa.gov/missions/cadre},
  accessed on 2023-08-24.

\bibitem{arches_dlr_2020}
M.~Schuster, M.~Müller, S.~Brunner, H.~Lehner, P.~Lehner, R.~Sakagami,
  A.~Dömel, L.~Meyer, B.~Vodermayer, E.~Giubilato, M.~Vayugundla, J.~Reill,
  F.~Steidle, I.~von Bargen, K.~Bussmann, R.~Belder, P.~Lutz, W.~Stürzl,
  M.~Smíšek, M.~Maier, S.~Stoneman, A.~Prince, B.~Rebele, M.~Durner,
  E.~Staudinger, S.~Zhang, R.~Pöhlmann, E.~Bischoff, C.~Braun, S.~Schröder,
  E.~Dietz, S.~Frohmann, A.~Börner, H.~Hübers, B.~Foing, R.~Triebel,
  A.~Albu-Schäffer, and A.~Wedler, ``The {ARCHES} space-analogue demonstration
  mission: Towards heterogeneous teams of autonomous robots for collaborative
  scientific sampling in planetary exploration,'' \emph{IEEE Robotics and
  Automation Letters (RA-L)}, vol.~5, no.~4, pp. 5315--5322, 2020.

\bibitem{slam_survey_leonard_2016}
C.~Cadena, L.~Carlone, H.~Carrillo, Y.~Latif, D.~Scaramuzza, J.~Neira, I.~Reid,
  and J.~Leonard, ``Past, present, and future of simultaneous localization and
  mapping: Toward the robust-perception age,'' \emph{IEEE Trans. on Robotics
  (TRO)}, vol.~32, no.~6, pp. 1309--1332, 2016.

\bibitem{distsurvey_halsted_2021}
T.~Halsted, O.~Shorinwa, J.~Yu, and M.~Schwager, ``A survey of distributed
  optimization methods for multi-robot systems,'' arXiv preprint, \unskip\space
  {arXiv}:2103.12840 [cs.{RO}], 2021.

\bibitem{choudhary_admm_2015}
S.~Choudhary, L.~Carlone, H.~Christensen, and F.~Dellaert, ``Exactly sparse
  memory efficient {SLAM} using the multi-block alternating direction method of
  multipliers,'' in \emph{Proc. IEEE/RSJ Intl. Conf. on Intelligent Robots and
  Systems (IROS)}, Hamburg, {DE}, Oct. 2015, pp. 1349--1356.

\bibitem{mbadmm_bad_chen_2016}
C.~Chen, B.~He, Y.~Ye, and X.~Yuan, ``The direct extension of {ADMM} for
  multi-block convex minimization problems is not necessarily convergent,''
  \emph{Mathematical Programming}, vol. 155, no. 1-2, pp. 57--79, 2016.

\bibitem{ddfsam_cunningham_2010}
A.~Cunningham, M.~Paluri, and F.~Dellaert, ``{DDF-SAM}: Fully distributed slam
  using constrained factor graphs,'' in \emph{Proc. IEEE/RSJ Intl. Conf. on
  Intelligent Robots and Systems (IROS)}, Taipei, {TW}, Oct. 2010, pp.
  3025--3030.

\bibitem{ddfsam2_cunningham_2013}
A.~Cunningham, V.~Indelman, and F.~Dellaert, ``{DDF-SAM 2.0}: Consistent
  distributed smoothing and mapping,'' in \emph{Proc. IEEE Intl. Conf. on
  Robotics and Automation (ICRA)}, Karlsruhe, {DE}, May 2013, pp. 5220--5227.

\bibitem{robot_web_murai_2022}
R.~Murai, J.~Ortiz, S.~Saeedi, P.~Kelly, and A.~Davison, ``A robot web for
  distributed many-device localisation,'' arXiv preprint, \unskip\space
  arXiv:2202.03314[cs.{RO}], 2022.

\bibitem{loopy_bp_murphy_1999}
K.~Murphy, Y.~Weiss, and M.~Jordan, ``Loopy belief propagation for approximate
  inference: An empirical study,'' in \emph{Proc. Fifteenth Conference on
  Uncertainty in Artificial Intelligence}, Stockholm, {SE}, 1999, p. 467–475.

\bibitem{dgs_choudhary_2017}
S.~Choudhary, L.~Carlone, C.~Nieto, J.~Rogers, H.~Christensen, and F.~Dellaert,
  ``Distributed mapping with privacy and communication constraints: Lightweight
  algorithms and object-based models,'' \emph{Intl. J. of Robotics Research
  (IJRR)}, vol.~36, no.~12, pp. 1286--1311, 2017.

\bibitem{chordal_init_carlone_2015}
L.~Carlone, R.~Tron, K.~Daniilidis, and F.~Dellaert, ``Initialization
  techniques for {3D} {SLAM}: A survey on rotation estimation and its use in
  pose graph optimization,'' in \emph{Proc. IEEE Intl. Conf. on Robotics and
  Automation (ICRA)}, Seattle, {USA}, 2015, pp. 4597--4604.

\bibitem{geod_cristofalo_2020}
E.~Cristofalo, E.~Montijano, and M.~Schwager, ``{Geod}: Consensus-based
  geodesic distributed pose graph optimization,'' arXiv preprint, \unskip\space
  {arXiv}:2010.00156 [cs.{RO}], 2020.

\bibitem{dc2pgo_tian_2021}
Y.~Tian, K.~Khosoussi, D.~Rosen, and J.~How, ``Distributed certifiably correct
  pose-graph optimization,'' \emph{IEEE Trans. on Robotics (TRO)}, vol.~37,
  no.~6, pp. 2137--2156, 2021.

\bibitem{asapp_tian_2020}
Y.~Tian, A.~Koppel, A.~Bedi, and J.~How, ``Asynchronous and parallel
  distributed pose graph optimization,'' \emph{IEEE Robotics and Automation
  Letters (RA-L)}, vol.~5, no.~4, pp. 5819--5826, 2020.

\bibitem{cadmm_mateos_2010}
G.~Mateos, J.~Bazerque, and G.~Giannakis, ``Distributed sparse linear
  regression,'' \emph{IEEE Transactions on Signal Processing}, vol.~58, no.~10,
  pp. 5262--5276, 2010.

\bibitem{boyd_distributed_2010}
S.~Boyd, ``Distributed optimization and statistical learning via the
  alternating direction method of multipliers,'' \emph{Foundations and Trends
  in Machine Learning}, vol.~3, no.~1, pp. 1--122, 2010.

\bibitem{cadmm_lin_converge_shi_2014}
W.~Shi, Q.~Ling, K.~Yuan, G.~Wu, and W.~Yin, ``On the linear convergence of the
  {ADMM} in decentralized consensus optimization,'' \emph{IEEE Transactions on
  Signal Processing}, vol.~62, no.~7, pp. 1750--1761, 2014.

\bibitem{sova_shorinwa_2020}
O.~Shorinwa, T.~Halsted, and M.~Schwager, ``Scalable distributed optimization
  with separable variables in multi-agent networks,'' in \emph{Proc. American
  Control Conference (ACC)}, Denver, {USA}, Jul. 2020, pp. 3619--3626.

\bibitem{thesis_wei_2014}
E.~Wei, ``Distributed optimization and market analysis of networked systems,''
  PhD thesis, Massachusetts Institute of Technology, Boston, MA, September
  2014.

\bibitem{o1k_convergence_wei_2013}
E.~Wei and A.~Ozdaglar, ``On the {O(1/k)} convergence of asynchronous
  distributed alternating direction method of multipliers,'' in \emph{Proc.
  IEEE Global Conference on Signal and Information Processing}, Austin, {USA},
  2013, pp. 551--554.

\bibitem{factor_graphs_for_robot_perception}
F.~Dellaert and M.~Kaess, ``Factor graphs for robot perception,''
  \emph{Foundations and Trends in Robotics (FNT)}, vol.~6, no. 1-2, pp. 1--139,
  2017.

\bibitem{admm_nonconvex_converge_hong_2016}
M.~Hong, Z.~Luo, and M.~Razaviyayn, ``Convergence analysis of alternating
  direction method of multipliers for a family of nonconvex problems,''
  \emph{SIAM Journal on Optimization}, vol.~26, no.~1, pp. 337--364, 2016.

\bibitem{dellaert_gtsam_tech_report_2012}
F.~Dellaert, ``Factor graphs and {GTSAM}: A hands-on introduction,'' Georgia
  Institute of Technology, Technical Report, 2012.

\bibitem{interpolation_survey_haarbach_2018}
A.~Haarbach, T.~Birdal, and S.~Ilic, ``Survey of higher order rigid body motion
  interpolation methods for keyframe animation and continuous-time trajectory
  estimation,'' in \emph{Proc. International Conference on 3D Vision (3DV)},
  Verona, {IT}, 2018, pp. 381--389.

\bibitem{madamm_kovnatsky_2016}
A.~Kovnatsky, K.~Glashoff, and M.~Bronstein, ``{MADMM}: a generic algorithm for
  non-smooth optimization on manifolds,'' in \emph{Proc. Eur. Conf. on Computer
  Vision (ECCV)}, Amsterdam, {NL}, 2016, pp. 680--696.

\bibitem{microlie_sola_2021}
J.~Solà, J.~Deray, and D.~Atchuthan, ``A micro lie theory for state estimation
  in robotics,'' arXiv, \unskip\space 1812.01537 [cs.{RO}], 2021.

\bibitem{isam_kaess_2007}
M.~Kaess, A.~Ranganathan, and F.~Dellaert, ``{iSAM}: Fast incremental smoothing
  and mapping with efficient data association,'' in \emph{Proc. IEEE Intl.
  Conf. on Robotics and Automation (ICRA)}, Rome, Italy, Apr. 2007, pp.
  1670--1677.

\end{thebibliography}
